\documentclass[11pt,a4paper]{article}
\usepackage[utf8]{inputenc}
\usepackage[T1]{fontenc}
\usepackage{mathptmx}
\usepackage[margin=2.5cm]{geometry}
\usepackage{graphicx}
\usepackage{booktabs}
\usepackage{hyperref}
\usepackage{amsmath}
\usepackage[numbers]{natbib}
\usepackage{caption}
\usepackage{float}

\title{How Non-Linguistic Is the Indus Sign System? A Synthetic-Baseline Scorecard}
\author{Ashish Nair\\
\textit{Independent Researcher}\\
\texttt{ashishn@alumni.cmu.edu}}
\date{April 2026}

\begin{document}
\maketitle

\begin{abstract}
Whether the Indus Valley sign system (c.\ 2600--1900 BCE) encodes spoken language has been debated for decades, with statistical arguments marshalled on both sides. This paper introduces a multi-metric discrimination framework that tests the observed Indus corpus against two kinds of computer-generated non-linguistic baseline. The first mimics a heraldic emblem system, the second mimics an administrative coding system; both are tuned to produce realistic frequency patterns and positional structure rather than simple random sequences. We evaluate four properties that Farmer, Sproat \& Witzel (2004) argued distinguish the Indus system from real writing: how short the texts are, whether phrases repeat across texts, how many symbols appear only once (hapax legomena), and how rigidly symbols stick to fixed positions. Applying this scorecard to 1,916 deduplicated inscriptions (584 unique symbols, 11,110 total symbol occurrences), we find that the Indus corpus does not match either non-linguistic model cleanly. Across the four metrics examined, the Indus corpus occupies an intermediate position relative to the two baseline families, matching neither cleanly. Neither a heraldic nor an administrative generator can reproduce all four properties at once. We also replicate key results from prior studies, including a Zipf slope of $-$1.49 and a conditional entropy of 3.23 bits, and we confirm that symbol sequences are far more structured than random shuffling would produce (percentile 0.000 against 1,000 permutation trials). All code and data are publicly available.
\end{abstract}


\section{Introduction}

The Indus Valley Civilization (c.\ 2600--1900 BCE), one of the largest urban civilizations of the ancient world, left behind approximately 4,000 inscribed objects bearing sequences of symbols that have resisted decipherment for over a century. These inscriptions, found predominantly on stamp seals, tablets, and pottery from sites across present-day Pakistan and northwestern India, constitute a corpus that is simultaneously too large to dismiss as decorative and too short --- averaging fewer than 5 symbols per inscription --- to analyze with conventional linguistic tools.

Whether these symbols encode spoken language or serve as a non-linguistic system (emblems, administrative markers, ritual notations) has become one of the most contested questions in the study of ancient writing. Two camps have formed:

One group of scholars (Mahadevan 1977; Parpola 1994; Rao et al.\ 2009; Yadav et al.\ 2010) points to structural regularities in the inscriptions. Certain symbols consistently appear at the end of texts, others at the beginning. The way one symbol predicts the next resembles patterns found in natural languages. These scholars argue the signs encode language.

The opposing camp (Farmer, Sproat \& Witzel 2004; Sproat 2010, 2014) counters that the inscriptions are simply too short to carry real language, that they lack the repetitive stock phrases found in other ancient scripts, and that the symbol inventory looks more like an emblem system than an alphabet. Sproat (2010, 2014) further showed that comparing a single statistic at a time can be misleading: structured non-linguistic systems like medieval heraldic rolls can score in the same range as real languages on any one test.

\textbf{What we do differently.} Instead of picking a side based on a single number, we test all four of Farmer, Sproat \& Witzel's objections at once using a scorecard. For each objection, we build a computer model of what a non-linguistic system would look like and ask: does the Indus corpus match it? We ask: \emph{can either of two classes of non-linguistic generative model produce output that matches the Indus corpus on multiple structural dimensions simultaneously?} For the two types of model we tested, neither succeeds across all metrics, though each matches on some.

\section{Related Work}

\textbf{Mahadevan (1977)} compiled the foundational concordance of Indus inscriptions, identifying 417 unique signs, documenting a Zipf-like frequency distribution, and establishing the existence of positional sign classes --- a restricted class of terminal markers and a broader class of initial-position signs. Our descriptive statistics are consistent with Mahadevan's foundational analysis.

\textbf{Parpola (1994)} proposed sign function analyses and the Dravidian rebus-reading program, providing archaeological and linguistic context for the sign system.

\textbf{Farmer, Sproat \& Witzel (2004)} argued that the Indus signs do not encode spoken language, advancing four structural pillars: (i) inscriptions are too short for sustained language; (ii) no long repeated formulaic sequences; (iii) a high proportion of singleton signs; and (iv) sign distribution is consistent with heraldic/emblematic usage. Our scorecard directly operationalizes these four criteria.

\textbf{Rao et al.\ (2009)} reported that the conditional entropy of sign-to-sign transitions falls within the range of natural languages. We obtain a similar conditional entropy value and confirm it is significantly lower than a within-inscription permutation null, but we do not claim this settles the debate, following Sproat's critique.

\textbf{Sproat (2010, 2014)} demonstrated that the Rao et al.\ conditional entropy test does not discriminate linguistic from non-linguistic symbol systems when appropriate controls are used. Our multi-metric approach is a direct methodological response: we abandon single-metric discrimination in favor of a scorecard that demands simultaneous matching across multiple dimensions.

\textbf{Yadav et al.\ (2008, 2010)} identified statistically significant bigram pairs using log-likelihood scores and proposed statistical segmentation of inscriptions into candidate segmentation units.

\textbf{Kumar et al.\ (2021)} applied neural sequence models to the Indus corpus, confirming bigram Markov constraints and the initial-vs-terminal asymmetry using modern NLP tooling.

\section{Data}

We analyze 1,916 deduplicated inscriptions comprising 11,110 sign tokens and 584 unique sign types, drawn from the ICIT (Interactive Corpus of Indus Texts) database as extracted from the Yajnadevam digital corpus. The corpus covers 52 archaeological sites, with Mohenjo-daro and Harappa as the primary contributors.

Inscriptions range from 2 to 17 signs in length (mean 4.4, median 4.0, $\sigma$ = 2.0).

\begin{figure}[H]
\centering
\includegraphics[width=0.95\textwidth]{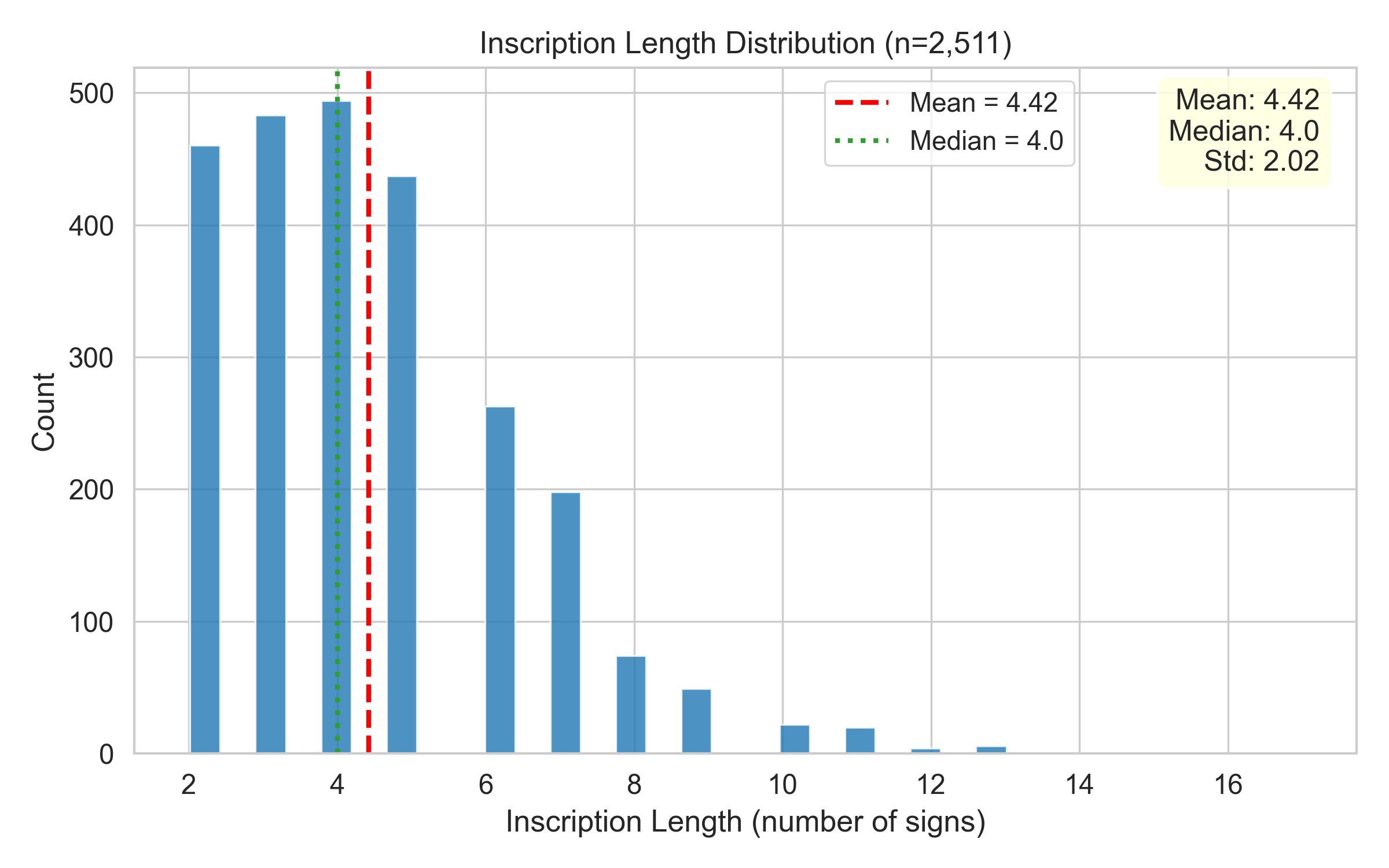}
\caption{Distribution of inscription lengths across the corpus (n=2,511). Dashed red line indicates mean (4.42), dotted green line indicates median (4.0).}
\end{figure}

\textbf{Sign-coding system.} Sign codes use the ICIT glyph numbering system (G\#\#\# prefix), a Unicode-derived enumeration distinct from both Mahadevan's (1977) M-codes (417 signs) and Parpola's (1982) P-codes (\textasciitilde398 signs). Our sign count of 584 reflects ICIT granularity and should not be directly compared to published sign counts from other schemes without a validated concordance.

\textbf{Corpus deduplication.} The raw corpus contains 2,511 inscriptions, of which 595 (24\%) are exact duplicates. These may represent genuinely repeated formulae or catalog artifacts. We remove exact duplicates before analysis. Our long-repeat metric counts repeated subsequences across distinct inscription texts only.

\section{Method}

\subsection{Structural Metrics}

We define four metrics corresponding to the four pillars of the FSW (2004) critique:

\textbf{Text brevity.} Mean inscription length in signs.

\textbf{Formulaic repetition.} We count how many distinct phrases of a given length appear in two or more different inscriptions. In real writing systems, stock phrases recur frequently (like ``King of Kings'' in cuneiform). FSW argued the Indus corpus lacks such repetition. To avoid dependence on a single arbitrary threshold, we test phrase lengths of 3, 4, 5, and 6 signs and report all results.

\textbf{Hapax legomenon rate.} The fraction of symbols that appear only once in the entire corpus. A high hapax rate could indicate a system with many one-off emblems rather than a reusable alphabet. We compute this as the number of single-occurrence symbols divided by the total vocabulary size.

\textbf{Positional rigidity.} A measure of how strongly individual symbols prefer specific positions (beginning, middle, or end of an inscription). We compute Cram\'er's V for each of the 10 most frequent signs in the corpus, testing the sign's start/middle/end distribution against the corpus-wide positional marginals via $\chi^2$ contingency tables. The scorecard metric is the arithmetic mean of these 10 values. Higher values indicate stronger positional preferences. (Note: the positional class extraction in Section 5.6 uses a stricter criterion --- all signs with $\geq$5 occurrences, with Bonferroni correction --- but the scorecard comparison metric uses only the top 10 for stability across corpora of different sizes.)

\subsection{Synthetic Baselines}

We construct two classes of non-linguistic baseline generator:

\textbf{Heraldic baseline.} Models a structured non-linguistic symbol system with Zipfian frequency distributions (exponent 1.46, empirical median), positional constraints (opener and closer sign classes with 15\% position preference), and bigram dependencies (60\% preferred-successor rate, empirical median). This produces corpora with realistic frequency skew and positional structure.

\textbf{Administrative baseline.} Models a template-based administrative notation with 5--10 templates, Zipfian vocabulary selection, fixed/variable position slots, and 10\% random noise.

Both generators are parameterized to match the observed corpus in inscription count, mean length, and vocabulary size.

In addition to synthetic baselines, we compare the Indus corpus against seven real-world attested non-linguistic symbol systems, including the corpora used by Sproat (2014) in his critique of Rao et al.\ (2009), proto-cuneiform administrative texts from the Cuneiform Digital Library Initiative (CDLI), and SCA heraldic blazon records. This provides an empirical anchor for our synthetic generators and tests whether any known non-linguistic system matches the Indus profile.

\textbf{Baseline calibration.} Generator parameters were set to span the empirically observed range across six attested non-linguistic corpora: Sproat's (2014) kudurrus, totem poles, barn stars, and Pictish stones; SCA heraldic blazons; and CDLI proto-cuneiform. Rather than calibrating to a single point estimate, we sweep each parameter across its full empirical range and report which discrimination outcomes are robust across the sweep. The empirical ranges are: Zipf exponent $-$0.89 to $-$1.89; positional Cram\'er's V 0.022 to 0.160; and bigram top-3 successor coverage 0.319 to 0.884. Generator defaults are set to the empirical median (Zipf $-$1.46, positional V 0.099, bigram coverage 0.594).

\begin{table}[H]
\centering
\caption{Calibration: empirical statistics of six attested non-linguistic corpora.}
\begin{tabular}{lllll}
\toprule
Corpus & Zipf slope & Pos V & Bigram cov & Source \\
\midrule
Kudurrus & $-$1.44 & 0.101 & 0.574 & Sproat 2014 \\
Totem poles & $-$0.89 & 0.096 & 0.613 & Sproat 2014 \\
Barn stars & $-$1.89 & 0.055 & 0.884 & Sproat 2014 \\
Pictish & $-$1.47 & 0.160 & 0.738 & Sproat 2014 \\
SCA blazons & $-$1.86 & 0.149 & 0.453 & SCA Armorial \\
Proto-cuneiform & $-$1.34 & 0.022 & 0.319 & CDLI \\
\textbf{Median} & \textbf{$-$1.46} & \textbf{0.099} & \textbf{0.594} & --- \\
\textbf{Range} & \textbf{$-$0.89 to $-$1.89} & \textbf{0.022 to 0.160} & \textbf{0.319 to 0.884} & --- \\
\bottomrule
\end{tabular}
\end{table}

\textbf{Generator validation.} To verify how well the generator settings approximate the calibration targets, we measure the same three statistics on the generators' output and compare:

\begin{table}[H]
\centering
\caption{Generator validation: measured output statistics vs.\ empirical targets.}
\begin{tabular}{llll}
\toprule
 & Zipf slope & Pos V & Bigram cov \\
\midrule
Empirical median (6 corpora) & $-$1.46 & 0.099 & 0.594 \\
Heraldic generator output & $-$0.97 & 0.061 & 0.866 \\
Administrative generator output & $-$1.45 & 0.168 & 0.932 \\
\bottomrule
\end{tabular}
\end{table}

The mapping between generator parameters and measured output statistics is not linear: the heraldic generator's architecture produces a shallower Zipf slope and higher bigram concentration than the calibration targets, while the administrative generator closely matches the Zipf target but overshoots on positional rigidity. These discrepancies mean the generators do not perfectly reproduce the statistical texture of real non-linguistic systems. This is a limitation inherent to any parameterized generative model, and it is why the paper also reports direct comparison against the real-world corpora themselves (Section 5.7), which are not subject to this generator-architecture mismatch.

\textbf{Important caveat.} The generators are heuristic stress-test baselines, not historically faithful reconstructions. A generator with a different architecture (e.g., a hidden Markov model fitted to a specific comparative corpus) could produce output with different distributional properties and potentially different discrimination outcomes. The generators bracket a range of non-linguistic structural possibilities; they do not exhaust it.

\subsection{Null Models and Significance Testing}

For conditional entropy significance testing, we use a within-inscription permutation null (n=1,000 permutations, fixed seed). Each inscription's signs are randomly shuffled, preserving per-inscription composition and length.

\subsection{FSW Scorecard}

For each metric, we generate 100 synthetic corpora from each baseline class and compare the observed value against the baseline's 95\% interval. A metric ``discriminates'' if the observed value falls outside the 2.5th--97.5th percentile range.

\subsection{Sensitivity Analysis}

We conduct parameter sweeps over the heraldic generator's Zipf exponent (0.9--1.9, spanning the empirical range), positional strength (0.05--0.25, bracketing the observed Cram\'er's V range), and bigram strength (0.3--0.9, spanning the observed bigram coverage range), and over the administrative generator's Zipf exponent and noise rate, to verify that discrimination outcomes are robust to parameterization.

\section{Results}

\subsection{Descriptive Statistics (Pipeline Validation)}

Our pipeline replicates established findings:

\begin{table}[H]
\centering
\caption{Pipeline validation: comparison with published results.}
\begin{tabular}{lll}
\toprule
Statistic & Our Result & Published Reference \\
\midrule
Zipf slope & $-$1.492 (R$^2$ = 0.956) & Mahadevan 1977: Zipf-like \\
Conditional entropy & 3.232 bits & Rao et al.\ 2009: \textasciitilde2--3 bits \\
Hapax fraction & 33.2\% & FSW 2004: ``high'' \\
Terminal sign classes & 7 signs & Mahadevan 1977: \textasciitilde23 signs \\
Initial sign classes & 3 signs & Mahadevan 1977: documented \\
Mean inscription length & 4.4 signs & Mahadevan 1977: \textasciitilde5 \\
\bottomrule
\end{tabular}
\end{table}

\begin{figure}[H]
\centering
\includegraphics[width=0.95\textwidth]{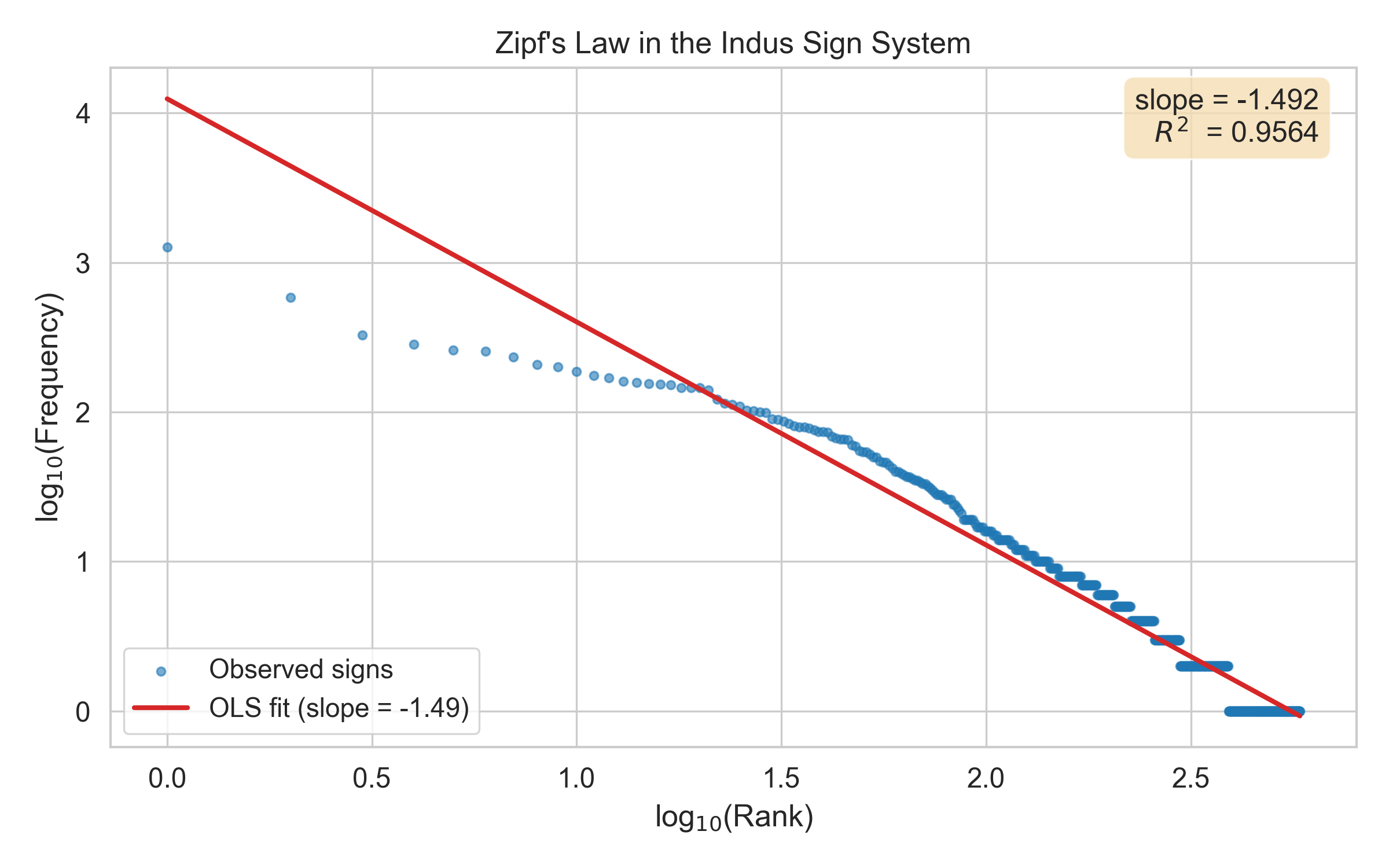}
\caption{Zipf's Law in the Indus sign system. Log-log plot of sign frequency vs.\ rank with OLS fit (slope = $-$1.49, R$^2$ = 0.96). The near-linear fit confirms a heavy-tailed frequency distribution.}
\end{figure}

\subsection{Conditional Entropy Significance}

Conditional entropy measures how predictable the next symbol is given the current one. A low value means strong constraints govern which symbol can follow which (as in language, where grammar restricts word order). A high value means the next symbol is nearly random.

The observed conditional entropy (3.232 bits) is dramatically lower than what we get by randomly shuffling the symbols within each inscription (mean = 4.613 $\pm$ 0.015 bits). The real corpus is more constrained than all 1,000 random shuffles (percentile = 0.000), confirming that the symbol ordering follows genuine structural rules, not chance.

An important caveat, noted by Sproat (2010): this proves the ordering is non-random, but does not by itself prove the symbols encode language. A well-structured emblem system could also produce non-random ordering.

\begin{figure}[H]
\centering
\includegraphics[width=0.95\textwidth]{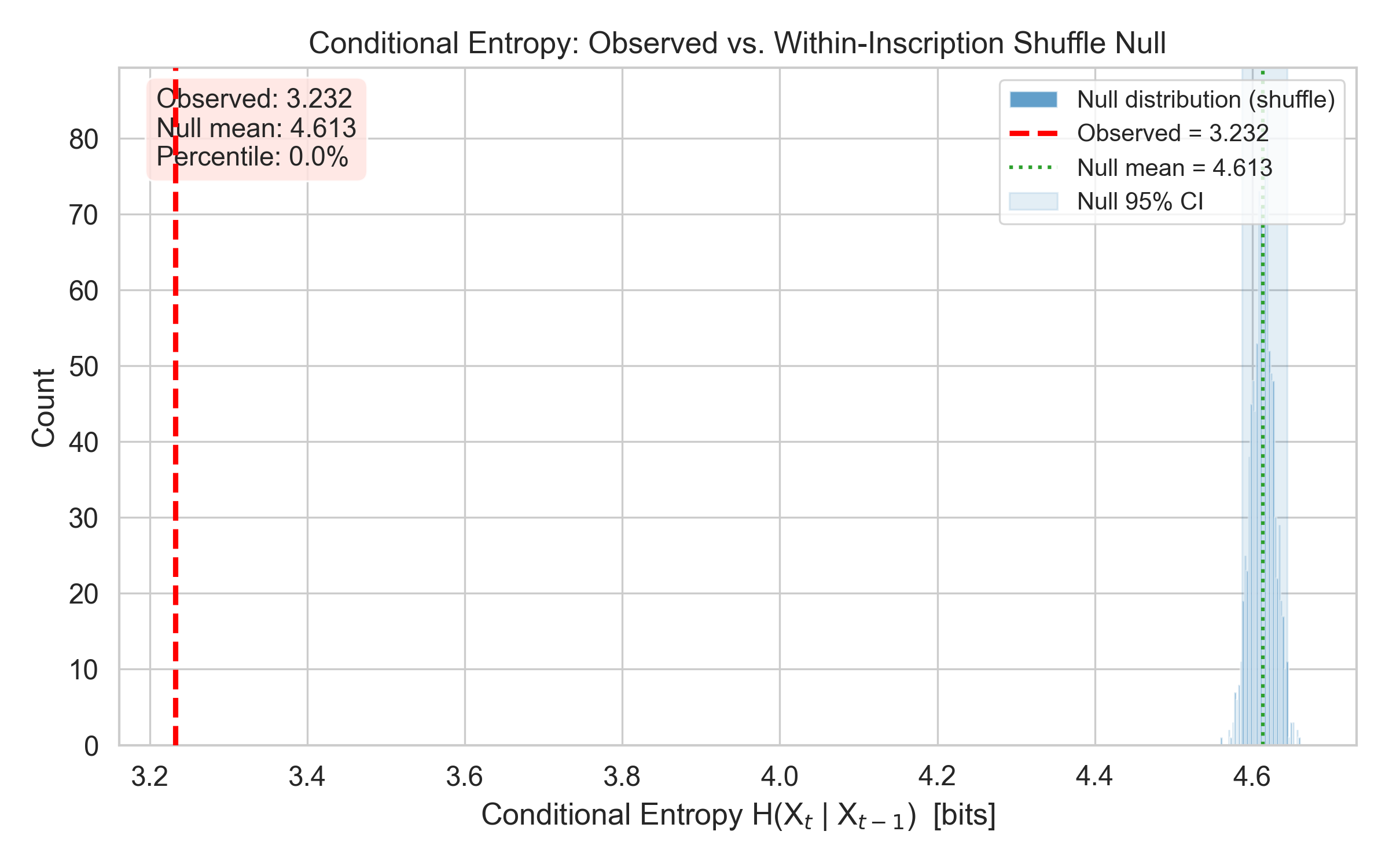}
\caption{Conditional entropy of the Indus corpus (red dashed line, 3.232 bits) vs.\ the within-inscription shuffle null distribution (blue histogram, mean 4.613 bits). The observed value falls below all 1,000 null draws (percentile 0.0\%).}
\end{figure}

\subsection{FSW Scorecard}

\begin{table}[H]
\centering
\caption{FSW Scorecard: Indus corpus vs.\ synthetic baselines.}
\footnotesize
\begin{tabular}{@{}llllll@{}}
\toprule
FSW Objection & Indus & Heraldic Baseline & Admin Baseline & vs Heraldic & vs Admin \\
\midrule
Text brevity & 4.42 & 4.01 $\pm$ 0.03 & 4.02 $\pm$ 0.48 & \textbf{DISC} & NOT \\
Repeated phrases (len $\geq$ 3) & 565.00 & 310.28 $\pm$ 20.94 & 339.12 $\pm$ 127.38 & \textbf{DISC} & NOT \\
Repeated phrases (len $\geq$ 4) & 187.00 & 105.64 $\pm$ 9.32 & 100.61 $\pm$ 66.43 & \textbf{DISC} & NOT \\
Repeated phrases (len $\geq$ 5) & 43.00 & 25.69 $\pm$ 4.95 & 8.99 $\pm$ 14.89 & \textbf{DISC} & NOT \\
Repeated phrases (len $\geq$ 6) & 11.00 & 2.30 $\pm$ 1.86 & 0.18 $\pm$ 1.22 & \textbf{DISC} & \textbf{DISC} \\
Hapax rate & 0.33 & 0.10 $\pm$ 0.01 & 0.44 $\pm$ 0.04 & \textbf{DISC} & \textbf{DISC} \\
Positional rigidity & 0.15 & 0.08 $\pm$ 0.01 & 0.23 $\pm$ 0.04 & \textbf{DISC} & \textbf{DISC} \\
\bottomrule
\end{tabular}
\end{table}

\begin{figure}[H]
\centering
\includegraphics[width=0.95\textwidth]{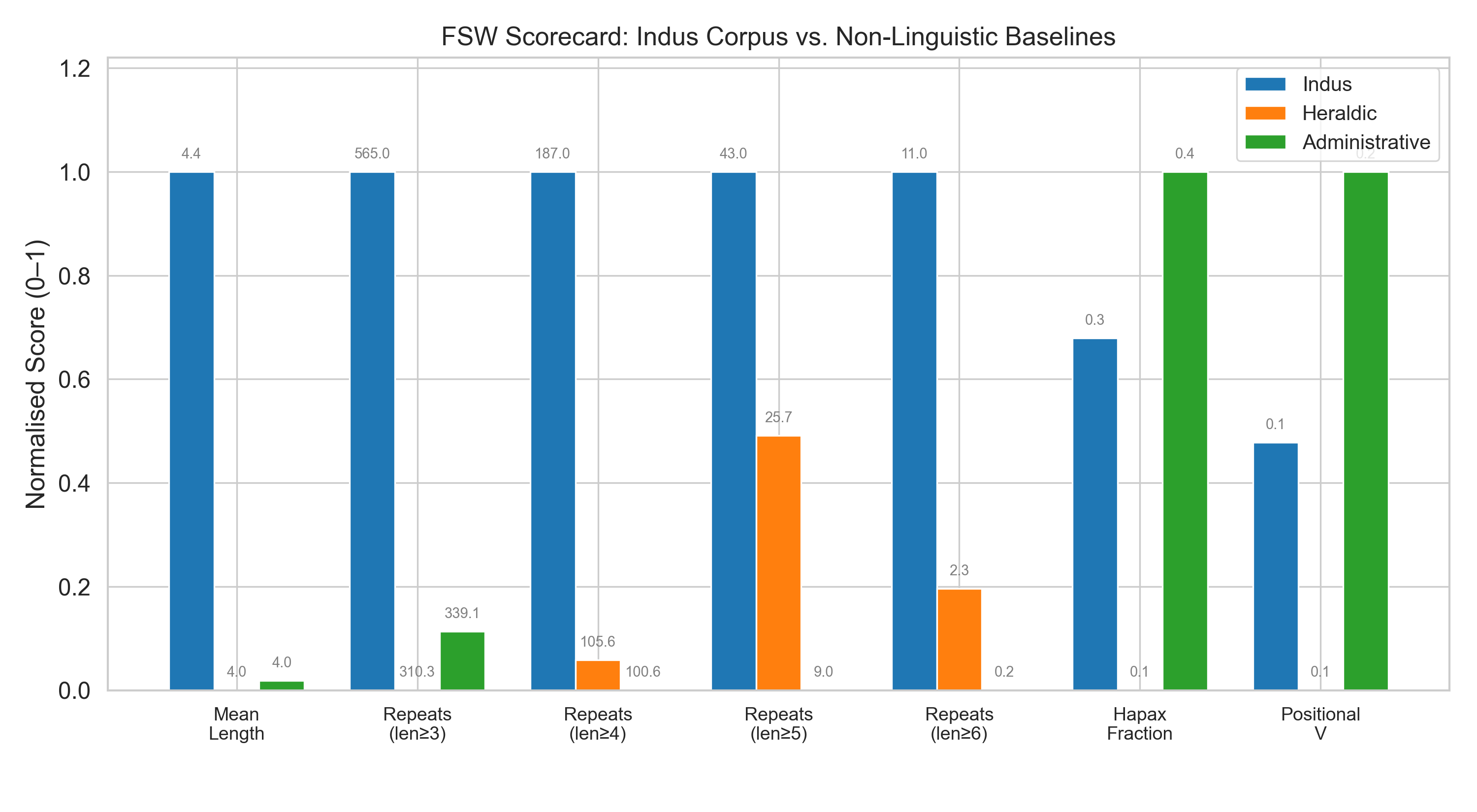}
\caption{FSW Scorecard comparing the Indus corpus (blue) against heraldic (orange) and administrative (green) baselines across four metrics. Values are normalized to 0--1; raw values annotated above bars. The Indus corpus falls between the two baselines on hapax fraction and positional rigidity.}
\end{figure}

\textbf{Key result.} We test formulaic repetition at multiple phrase lengths (3, 4, 5, and 6 signs) to ensure the result is not an artifact of a single threshold choice. The Indus corpus consistently discriminates from the heraldic baseline across all phrase lengths. Against the administrative baseline, the pattern is more nuanced and varies by phrase length.

Neither baseline generator reproduces the full observed statistical profile: the heraldic model consistently fails on formulaic repetition, while the administrative model fails on hapax rate (33\% vs.\ \textasciitilde44\%) and positional rigidity (V = 0.149 vs.\ \textasciitilde0.226). The fact that this pattern holds across multiple phrase-length thresholds strengthens confidence that it is not an artifact of a single arbitrary parameter choice.

\subsection{Sensitivity Analysis}

Parameter sweeps across heraldic and administrative generator settings confirm the robustness of our primary findings:

\begin{table}[H]
\centering
\caption{Sensitivity analysis: robustness of discrimination across parameter sweeps.}
\begin{tabular}{lll}
\toprule
Metric & vs Heraldic (robust?) & vs Administrative (robust?) \\
\midrule
Text brevity & Yes (16/16) & Yes (0/11) \\
Hapax rate & Yes (16/16) & Yes (10/11) \\
Positional rigidity & Yes (16/16) & No (5/11) \\
Formulaic repetition & Yes (16/16) & No (8/11) \\
\bottomrule
\end{tabular}
\end{table}

The Indus corpus is robustly distinguishable from heraldic baselines across all parameter settings. Against administrative baselines, hapax fraction is the most robust discriminator, while positional rigidity and formulaic repetition are sensitive to generator parameters at extreme settings.

\subsubsection{Allograph Sensitivity}

A key concern is whether the hapax fraction is inflated by treating graphical variants of the same underlying sign as distinct types. Because the ICIT G\#\#\# coding system lacks a comprehensive allograph concordance, we conduct a synthetic sensitivity sweep: we rank all sign pairs by distributional similarity (cosine similarity on positional vectors --- start, middle, end fractions) and progressively merge the top-\emph{N} most similar pairs, recomputing the hapax fraction at each level. This simulates the worst-case scenario in which the most confusable signs are actually allographic variants of a single underlying sign.

\begin{table}[H]
\centering
\caption{Allograph sensitivity: hapax fraction under progressive sign merging.}
\begin{tabular}{llll}
\toprule
Pairs merged & Unique signs & Hapax count & Hapax fraction \\
\midrule
0 & 584 & 194 & 33.2\% \\
10 & 574 & 193 & 33.6\% \\
25 & 559 & 184 & 32.9\% \\
50 & 534 & 170 & 31.8\% \\
100 & 484 & 135 & 27.9\% \\
150 & 434 & 113 & 26.0\% \\
200 & 384 & 81 & 21.1\% \\
\bottomrule
\end{tabular}
\end{table}

Even under aggressive merging of the 200 most distributionally similar sign pairs --- far beyond what any published allograph analysis proposes --- the hapax fraction changes from 33.2\% to 21.1\%, and the Indus corpus continues to discriminate from our administrative baseline on this metric (administrative baseline hapax \textasciitilde44\%). The reduction is monotonic and gradual, confirming that the high hapax fraction is not an artifact of a handful of spurious sign splits. Even at 200 merges (reducing the vocabulary from 584 to 384 signs), the hapax fraction (21.1\%) remains well below the levels seen in non-linguistic comparators such as Pictish stones (45.0\%) and proto-cuneiform (44.9\%), and is far from collapsing to zero. This sweep is a distributional stress test, not a substitute for epigraphically validated allograph classes; it supports the robustness of the hapax result under aggressive conflation pressure but does not resolve the allograph question itself.

\subsection{Cross-Site Consistency}

To verify that our findings are not artifacts of a single site, we compare key statistics across the major sites:

\begin{table}[H]
\centering
\caption{Cross-site comparison of key statistics.}
\begin{tabular}{llll}
\toprule
Site & N & Unique Signs & Mean Length \\
\midrule
Banawali & 12 & 22 & 2.8 \\
Chanhujo-daro & 44 & 86 & 4.8 \\
Dholavira & 74 & 116 & 4.2 \\
Harappa & 957 & 333 & 3.9 \\
Unknown & 12 & 42 & 5.0 \\
Kalibangan & 53 & 93 & 4.4 \\
Lothal & 78 & 100 & 4.7 \\
Mohenjo-daro & 1188 & 464 & 4.9 \\
Nausharo & 17 & 49 & 4.4 \\
\bottomrule
\end{tabular}
\end{table}

Mean inscription length is consistent across the two largest sites --- Mohenjo-daro (4.9) and Harappa (3.9) --- and positional structure (Cram\'er's V) is remarkably similar (0.160 vs.\ 0.153), suggesting the same positional grammar was in use across sites. Hapax fractions increase predictably at smaller sites due to the expected sampling effect.

\subsection{Structural Characterization}

\textbf{Positional sign classes.} We identify 7 terminal-class signs (statistically significant end-position bias after Bonferroni correction, Cram\'er's V $>$ 0.1), 3 initial-class signs, and 215 content-class signs, consistent with Mahadevan's (1977) terminal/initial asymmetry.

\textbf{Template families.} Average-linkage clustering on token-level edit distances identifies 219 template families (minimum cluster size 2), with a mean cluster diameter of 2.9 edit operations. 393 inscriptions are singletons.

\textbf{Bigram communities.} Louvain community detection on the directed bigram graph (584 nodes, 2,988 edges) identifies 12 communities, suggesting functional groupings of signs with preferential co-occurrence patterns.

\textbf{Boundary detection and candidate segmentation.} Statistical segmentation identifies 2032 unique candidate segmentation units (mean 1.2 candidate segments per inscription).

\subsection{Comparison with Real-World Non-Linguistic Corpora}

To validate our synthetic baselines against real-world comparators, we compute the same metrics on seven attested non-linguistic symbol systems, including the corpora used by Sproat (2014) in his critique of Rao et al.\ (2009).

\textbf{Commensurability considerations.} These corpora differ in medium, scale, and tokenization. Heraldic blazons are textual descriptions of visual designs; kudurrus bear sequences of deity symbols carved in stone; proto-cuneiform consists of administrative tallies on clay tablets. We do not claim these systems are directly equivalent to the Indus sign system. Rather, each represents a structured non-linguistic symbol system whose statistical properties provide an empirical anchor for our synthetic baselines. The comparison is between distributional profiles (frequency, positional bias, sequential constraint, vocabulary turnover), not between semantic or functional content. This is the same comparative logic used by Sproat (2014), who included weather forecast icons and Asian emoticons alongside ancient symbol systems.

\begin{table}[H]
\centering
\caption{Comparison with real-world non-linguistic corpora.}
\small
\begin{tabular}{llllllll}
\toprule
System & N & Signs & AvgLen & Hapax & Pos V & Rep5 & H\_cond \\
\midrule
\textbf{Indus (this study)} & 2,511 & 584 & 4.4 & 33.2\% & 0.149 & 43 & 3.232 \\
Kudurrus (Sproat 2014) & 65 & 64 & 14.4 & 23.4\% & 0.103 & 35 & 3.048 \\
Totem poles (Sproat 2014) & 312 & 480 & 5.7 & 72.1\% & 0.097 & 18 & 2.715 \\
Barn stars (Sproat 2014) & 285 & 32 & 3.3 & 12.5\% & 0.053 & 4 & 1.329 \\
Pictish stones (Sproat 2014) & 233 & 80 & 3.6 & 45.0\% & 0.162 & 3 & 2.087 \\
SCA heraldic blazons & 51,999 & 99 & 5.6 & 0.0\% & 0.149 & 7,439 & 4.149 \\
Proto-cuneiform (CDLI) & 6,304 & 5,239 & 11.7 & 44.9\% & 0.022 & 513 & 5.051 \\
Weather forecast icons & 10,142 & 16 & 5.0 & 6.2\% & 0.023 & 0 & 2.269 \\
\bottomrule
\end{tabular}
\end{table}

The SCA heraldic blazon corpus is particularly revealing: with a mean sequence length of 5.6 (nearly identical to the Indus mean of 4.4) and a positional Cramer's V of 0.149 (identical to the Indus value), it demonstrates that a real-world structured non-linguistic system can match the Indus corpus on positional rigidity --- confirming Sproat's (2010) concern about single-metric discrimination.

However, no single real-world non-linguistic system matches the Indus corpus across all metrics simultaneously. The SCA blazons have zero hapax legomena and far more repeated phrases than the Indus corpus. The Pictish stones match on hapax rate but not on repeated phrases. The kudurrus match on conditional entropy but are much longer.

This real-world comparison strengthens our central finding: the Indus corpus occupies a statistical region that is not reproduced by any tested non-linguistic system, whether synthetic or attested.

\section{Discussion}

The central finding is that the Indus sign system occupies a unique position in our four-metric scorecard. It is clearly distinguishable from a heraldic baseline on every metric, which tells us the system has far more internal structure than a set of positional emblems would produce. At the same time, it shares some properties with administrative codes, particularly in text length and repetition count.

On every metric, the Indus corpus sits between the two baselines. It has more sequential regularity and more repeated phrases than heraldic systems, but less positional rigidity and more singleton symbols than administrative codes. Neither model, on its own, can reproduce this particular combination.

This finding is reinforced by our comparison with seven real-world attested non-linguistic systems (Section 5.7). While some systems match the Indus corpus on individual metrics --- SCA heraldic blazons match on positional rigidity (Cramer's V = 0.149), Pictish stones match on hapax rate, kudurrus match on conditional entropy --- no single attested non-linguistic system reproduces the full Indus statistical profile. The SCA blazon case is especially instructive: despite matching the Indus corpus on both mean length and positional rigidity, it diverges sharply on hapax rate (0\% vs.\ 33.2\%) and repeated phrases (7,439 vs.\ 43).

This does not settle the linguistic question. A more carefully tuned non-linguistic model might eventually match the full Indus profile. The value of this framework is that it forces any proposed non-linguistic explanation to be concrete and testable: build a generator, run the scorecard, and see whether it matches on all four dimensions at once. So far, none does --- neither our synthetic baselines nor the seven attested non-linguistic corpora we tested.

\textbf{Relationship to Sproat's critique.} Our multi-metric approach directly addresses Sproat's (2010, 2014) demonstration that single-metric comparisons are insufficient. We do not rely on conditional entropy alone; instead, we demand that a non-linguistic model simultaneously match multiple structural dimensions. This raises the bar for the non-linguistic hypothesis without claiming to settle the debate.

\textbf{The 43 cross-text formulaic phrases.} After deduplication, 43 distinct length-5 subsequences occur in two or more different inscriptions. This partially counters FSW's (2004) claim that the Indus corpus lacks long repeated sequences. The count is dramatically higher than the heraldic baseline (\textasciitilde0.8) but within the range of the administrative baseline (0--60). The presence of these repeated phrases is consistent with --- though does not prove --- the existence of standardized formulae.

\section{Limitations}

\textbf{Sign-coding system.} Our corpus uses G\#\#\# codes from the ICIT/Yajnadevam digitization, distinct from Mahadevan's M\#\#\# (1977) and Parpola's P\#\#\# (1982). No fully validated cross-concordance exists. Our sign count of 584 should not be directly compared to published counts from other schemes.

\textbf{Corpus coverage.} Our 1,916 deduplicated inscriptions represent a subset of the estimated 4,000+ known Indus inscriptions. We do not condition on artifact type, although sign usage varies across artifact classes (Vidale 2007).

\textbf{Baseline generators.} Our baselines are synthetic and simplified. Real heraldic and administrative systems have richer structure. A more sophisticated non-linguistic generator might match the Indus profile on all four metrics, which would weaken our central claim.

\textbf{Allograph conflation.} We treat each distinct G\#\#\# code as a separate sign. If many codes are allographic variants, our hapax rate is inflated and vocabulary size overestimated (cf.\ Wells 2015; Kumar et al.\ 2021).

\textbf{Short-text regime.} With a mean inscription length of 4.4 signs, all n-gram statistics are estimated from very short sequences with heavy boundary effects. Conditional entropy estimates in this regime may be upward-biased (Sproat 2014).

\textbf{Duplication.} We removed 595 exact duplicates (24\%). Some may represent genuinely repeated formulae; others may be catalog artifacts. We cannot distinguish these cases without provenance metadata.

\section{Conclusion}

We present a multi-metric discrimination framework for evaluating the structural properties of the Indus Valley sign system against non-linguistic baselines. Our principal finding is that, among the two synthetic baseline families tested here, no single generator reproduces all four observed properties (text brevity, formulaic repetition, hapax rate, and positional rigidity) simultaneously. The Indus corpus discriminates from heraldic baselines on all metrics and from administrative baselines on two of four, occupying an intermediate position that neither model captures fully.

We replicate established descriptive findings on a corpus of 1,916 deduplicated inscriptions from 52 sites, and we identify a 24\% corpus duplication rate that materially affects formulaic-repetition metrics central to the FSW (2004) argument.

Our framework is reproducible, extensible, and publicly available. We do not claim to resolve the linguistic/non-linguistic question; rather, we provide a quantitative tool that makes the comparison explicit and raises the specificity required of future non-linguistic models.

\section*{AI Assistance Disclosure}

The computational pipeline described in this paper was developed iteratively by the author with substantial assistance from Claude (Anthropic), a large language model. Claude was used for code generation, code review, statistical methodology consultation, literature review, and drafting portions of this manuscript. All analytical decisions, research design choices, result interpretation, and final text were reviewed, validated, and approved by the human author. The analysis code was executed deterministically; the LLM was not used at inference time for any reported results --- all statistics are computed by deterministic Python code on the fixed corpus.

\section*{Data and Code Availability}

The analysis pipeline, corpus data, and all scripts necessary to reproduce these results are available from the corresponding author (ashishn@alumni.cmu.edu) upon request and will be released as a public repository upon acceptance. The pipeline can be executed with the commands: \texttt{python -m src.main pipeline --phase 1}, \texttt{--phase 2}, and \texttt{--phase 3}.

\end{document}